%% file: main.tex
\documentclass[10pt,journal,compsoc]{IEEEtran}

\usepackage{scalerel}
\usepackage{tikz}
\usetikzlibrary{svg.path}

\definecolor{orcidlogocol}{HTML}{A6CE39}
\tikzset{
  orcidlogo/.pic={
    \fill[orcidlogocol] svg{M256,128c0,70.7-57.3,128-128,128C57.3,256,0,198.7,0,128C0,57.3,57.3,0,128,0C198.7,0,256,57.3,256,128z};
    \fill[white] svg{M86.3,186.2H70.9V79.1h15.4v48.4V186.2z}
                 svg{M108.9,79.1h41.6c39.6,0,57,28.3,57,53.6c0,27.5-21.5,53.6-56.8,53.6h-41.8V79.1z M124.3,172.4h24.5c34.9,0,42.9-26.5,42.9-39.7c0-21.5-13.7-39.7-43.7-39.7h-23.7V172.4z}
                 svg{M88.7,56.8c0,5.5-4.5,10.1-10.1,10.1c-5.6,0-10.1-4.6-10.1-10.1c0-5.6,4.5-10.1,10.1-10.1C84.2,46.7,88.7,51.3,88.7,56.8z};
  }
}

\newcommand\orcidicon[1]{\href{https://orcid.org/#1}{\mbox{\scalerel*{
\begin{tikzpicture}[yscale=-1,transform shape]
\pic{orcidlogo};
\end{tikzpicture}
}{|}}}}

\usepackage{hyperref} 
\usepackage{orcidlink} 

\usepackage{dirtytalk}
\usepackage{acronym} 

\usepackage[
style=nature
]{biblatex}

\ifCLASSINFOpdf
\else
\fi
\begin{document}
%
\title{Advancing Earth Observation: A Survey on AI-Powered Image Processing in Satellites}
\author{Aidan Duggan$^1$,\orcidlink{0009-0009-5325-768X}
        Bruno Andrade$^1$,\orcidlink{0000-0003-2481-401X}
        and~Haithem Afli$^1$,\orcidlink{0000-0002-7449-4707}~\IEEEmembership{(Senior Member,~IEEE)}\\
        {\scriptsize $^1$Computer Science Department, Munster Technological University, Cork, T12 P928 Ireland}
\IEEEcompsocitemizethanks{\IEEEcompsocthanksitem This research was conducted with the financial support of the ADVANCE CRT Ph.D. Cohort under Grant Agreement No.18/CRT/6222 and the ADAPT SFI Research Centre at Munster Technological University. The ADAPT SFI Centre for Digital Media Technology is funded by Science Foundation Ireland through the SFI Research Centres Programme and is co-funded under the European Regional Development Fund (ERDF) through Grant 13/RC/2106\_P2}}
\IEEEtitleabstractindextext{%
\begin{abstract}
\normalsize
Advancements in technology and reduction in it's cost have led to a substantial growth in the quality \& quantity of imagery captured by Earth Observation (EO) satellites. This has presented a challenge to the efficacy of the traditional workflow of transmitting this imagery to Earth for processing. An approach to addressing this issue is to use pre-trained artificial intelligence models to process images on-board the satellite, but this is difficult given the constraints within a satellite's environment. This paper provides an up-to-date and thorough review of research related to image processing on-board Earth observation satellites. The significant constraints are detailed along with the latest strategies to mitigate them. 
\end{abstract}

\begin{IEEEkeywords}
\normalsize
smart satellite, deep learning, energy optimisation, satellite data analysis, edge computing
\end{IEEEkeywords}}

\maketitle

\IEEEdisplaynontitleabstractindextext

%
\IEEEpeerreviewmaketitle

\input{Sections/Introduction} 
\input{Sections/Background}  
\input{Sections/Constraints}  
\input{Sections/Solutions}  
\input{Sections/Conclusion}
\ifCLASSOPTIONcompsoc
  \section*{Acknowledgments}
\else
  \section*{Acknowledgment}
\fi

This research was conducted with the financial support of the ADVANCE CRT Ph.D. Cohort under Grant Agreement No.18/CRT/6222 and the ADAPT SFI Research Centre at Munster Technological University. The ADAPT SFI Centre for Digital Media Technology is funded by Science Foundation Ireland through the SFI Research Centres Programme and is co-funded under the European Regional Development Fund (ERDF) through Grant 13/RC/2106\_P2.

\ifCLASSOPTIONcaptionsoff
  \newpage
\fi


\label{Bibliography}

\printbibliography

\end{document}

%% file: Sections/Introduction.tex
\IEEEraisesectionheading{\section{Introduction}\label{sec:introduction}}

%
%
%
%
\IEEEPARstart{T}{here} has been a huge increase in the number of satellites being launched into space in recent years. As of November 2024, the satellite tracking website “Orbiting Now” \cite{orbit_now} lists over 10,500 active satellites in various Earth orbits. While the majority (80\%) are associated with communications 1052 of them are earth observation (EO) satellites generating thousands of terabytes of data every day. Relaying this volume of data to earth is not feasible through traditional radio frequency (RF) communication channels so other solutions have been investigated including processing the data on board the satellite where the data is produced. This solution is similar to the introduction of edge computing which is a model of distributing computation closer to the source of data, developed to manage the proliferation of devices connected to the internet commonly called Internet-of-Things(IoT). Machine Learning (ML) has been a key enabler to the success of edge computing.
\par An article published at the end of 2020 by Furano et al. \cite{furano2020towards} explores some of the compelling reasons that necessitate the deployment of ML on-board a satellite to do image processing. This includes the growth in the volume of data being produced by sensors not being matched by a corresponding increase in data download capacity, restricted power in smaller satellites to download large images and issues with ground station availability. The challenges are also pointed out including older hardware with insufficient resources, lack of on-board storage or working memory and limited availability of datasets required for model training and evaluation. The benefits identified were better responsiveness because of the reduction in data to be downloaded, improved results (including accuracy), bandwidth savings so less pressure on the communication channels and more flexibility because images can be selected or filtered depending on the application.
\par The concept of an Intelligent EO satellite system was introduced as far back as 2003 by Zhou and Kafatos \cite{zhou2003future} envisioning dynamic on-board integration of sensors, data processors and communication systems. Since then there have been multiple research efforts to try and achieve this vision. Prior reviews of the domain include Miralles et al. \cite{di2021machine} who did a very broad overview on research of ML in the EO domain for both on-ground and on-board operations. It covers topics from mission planning, telecommunications to on-board image processing. The same authors updated the review in early 2023 \cite{miralles2023critical} but there were very few additions to the on-board image processing section. While it presents a broad review across many aspects of ML in EO it does not have the depth to cover all of the literature of the specific domain of on-board image processing in detail. In early 2022, Zhang et al. \cite{zhang2022progress} reviewed developments in intelligent remote sensing satellite systems over the previous decade including sections on the remote sensing satellite platform, imaging payload and onboard processing systems. It also outlines the broader challenges of distributing the satellite data available to a large audience and documenting the improvements this brings regarding privacy protection, societal values, and laws regarding the sharing and distribution of data and information. In August 2023 Gardill et al. \cite{TowardsSE} presented a summary of findings related to developments and emerging applications of the space edge-computing and onboard artificial intelligence. It focuses on the space communication link and benchmarks deep-learning models on 2 commercial off the shelf edge processors on board the International Space Station(ISS). In early 2024 Thangavel et al. \cite{thangavel2024artificial} provided a review of AI for satellite operations, with a special focus on Distributed Satellite Systems. A great overview of the various spaceflight systems is given along with how satellite tasks can be made autonomous through the use of AI. There is very little information on processing imagery on-board a satellite but a strong argument for doing so is made, making reference to \cite{furano2020towards}.
\par To the best of our knowledge none of the existing reviews are dedicated to the deployment of ML onto EO satellites for the purpose of on-board processing. In such a dynamic domain there is a need for a deep, thorough and up to date review of all the developments to assist anyone trying to overcome the challenges. Specifically this survey addresses the following:
\begin{enumerate}
  \item What are the most challenging constraints to deploying pre-trained ML algorithms on-board an EO satellite to facilitate on-board image processing?
  \item What techniques are most effective in overcoming these constraints?
\end{enumerate}

The paper is organised as follows. Section \ref{Background} gives background information on satellites including how they are classified, the cost, typical on-board hardware and sensors and well as the data produced and how it in handled. Then a brief background is given on the evolution of Artificial Intelligence and Edge Computing.  Section \ref{ConstraintsondeployingMachineLearningtoSatellites} outlines the significant constraints to deploying AI on-board satellites in detail. Section \ref{Methodstomitigate} described various way to mitigate these constraints. 



%% file: Sections/Background.tex
\section{Background} \label{Background}
\subsection{Satellites}
In 1957 the first artificial satellite (called Sputnik) was launched by the Soviet Union. This was the culmination of many years research dating back over 3 centuries when Isaac Newton efforts attempted to explain the motion of the natural satellites such as the moon that orbit around planets \cite{Earth_satellite}. This started an industry that has resulted in over 10,500 satellites launched from 105 different countries. 
\subsubsection{Satellite Classification}
Satellites are primarily classified by mission 
\begin{itemize}
\item Communication - transmitting telecommunication signals such as TV, phone and Internet from one location on earth to another. These are traditionally located 35,785 km above the earth which is called geostationary or geosynchronous orbit (GEO), because they orbit at the same rotation speed as earth staying over the same point of earth. However this is rapidly changing so satellites at lower altitudes of 600-800 km (Low Earth Orbit - LEO) will dominate the future telecommunications market.
\item Earth Observation - As the name suggests this class of satellite is primarily used to take images of the earth used for various applications such as land/sea monitoring \& meteorology. They generally operate at LEO altitudes taking around 90 minutes (altitude dependant) to orbit the earth \cite{nasa_orbits}
\item Navigation / Global Positioning - provides position, tracking \& navigation abilities to devices on earth containing a electronic receiver using time signals transmitted along a line of sight from a combination of several satellites. They orbit at around 20,000 km (MEO - Medium Earth Orbit) taking around 12 hours to orbit the earth 
\end{itemize}
\subsubsection{Satellite Cost}
Historically satellites were custom built for specific tasks and cost in the order of hundreds of millions of dollars, so were only accessible to governments or very large companies. As new technologies, cheaper materials \& fuel types have emerged along with the miniaturisation and standardisation of electronic parts, satellites have got smaller and considerably cheaper and this trend is predicted to continue \cite{fox_2020}. Several commercial companies (eg. SpaceX in 2012) have entered the market and now offer a "Rideshare" model for small satellite companies to hitch a ride to space. A small satellite can now be built and deployed in orbit for as little as half a million dollars.  This has democratised the domain and made it accessible to a much wider population and given rise to opportunities for both the commercial and research sectors. 
\par
\subsubsection{Satellite Hardware}
\begin{figure}[ht]
\includegraphics{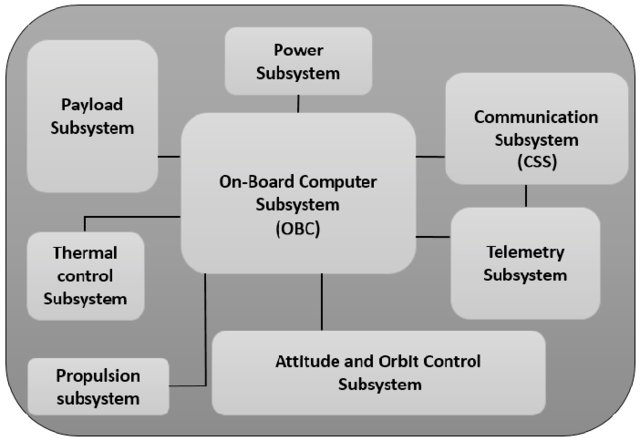}
\centering
\caption{LEO Satellite Block Diagram \cite{el2015new}}
\label{fig:LEO Satellite Block Diagram}
\end{figure}

A satellite is designed primarily to stay functional in space and communicate the data acquired from it's payload (sensors) to the ground station. Figure \ref{fig:LEO Satellite Block Diagram} shows the classical layout of a satellite with data handling being based on a central on-board computer (OBC) with minimal data processing. 
In 1999 \cite{heidt2000cubesat} Puig-Suari and Bob Twiggs developed the CubeSat specifications to promote  the design of small satellites intended for low Earth orbit (LEO). This form factor, of 10 cm cubes weighing no more than 2kg, has become the de-facto standard for small EO satellites (smallsats) design. Satellites are defined by the number of cubes (e.g. 6U is 6 cubes stacked in a 2 cubes side-by-side form).

\subsubsection{Satellite Sensors}
\begin{figure*}[ht]
\includegraphics[width=\textwidth]{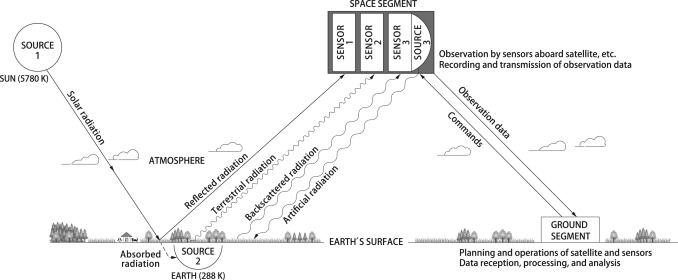}
\caption{Remote Sensing \cite{omnisci}}
\label{fig:remote-sensing}
\end{figure*}

There are 2 broad categories of earth observation sensors as shown in Figure~\ref{fig:remote-sensing}
\noindent \begin{itemize}
\item Passive - gathers radiation reflected or emitted from a target which have external sources. In Figure~\ref{fig:remote-sensing} there are 2 examples of external sources (1 - Sun, 2 - Earth). These are generally categorised as optical sensors operating at wavelengths typical for visible light(430–720 nm) or for infra-red(750–950 nm) and while they can get high resolution of small areas, they can only function during the day (i.e. need sunlight) and are affected by cloud and weather. Hyperspectral sensors collect information as a set of 'images'. Each image represents a narrow wavelength range of the electromagnetic spectrum. These 'images' are combined to form a three-dimensional hyperspectral data cube and can be used to detect certain minerals or objects if their spectral properties are known.
\item Active - Use an internal source which emits energy(radiation) directed at a target and a sensor gathers the reflected radiation. The difference in time between the emission and gathering is then used to identify properties of the target. These are more generally referred to as radar based (e.g. synthetic aperture radar (SAR)) operating at longer wavelengths (e.g.. microwaves) and can operate at any time and will not be impacted by weather. This is advantageous if constant coverage is required (e.g, for monitoring change at reduced spatial resolution). 
\end{itemize}	
\subsubsection{Satellite Data}
In 2017 an article by Intel \cite{data_2017} coined the term "Space Data: The Final Analytics Frontier" to bring peoples attention to the potential value of the data being obtained from satellites orbiting the earth thousands of miles above us. There has been a boom in the industry in the last 15 years due to the increase in the quality and quantity of imagery available. The global Small Satellite market size is projected to grow annually by over 20\% to \$6.6 by the end of 2024 and to over \$14 billion by 2028 \cite{small_satellite_global_market_2024}. The Copernicus is the European Union's Earth Observation Programme and provides freely available 20TB of data from it's fleet of 3 Sentinel Satellites to the community \cite{esch2018exploiting}. Commercial companies also offer over 150TB per day. The term "big Earth data" is now used to describe the massive earth observation datasets. With SpaceX planning to launch 12000 satellites by 2025 \cite{beall2020spotting} along with many others that data volume is set to grow exponentially. Imagery of the earth gathered from EO satellites are used for many purposes from monitoring weather, land usage, conflicts, traffic, transportation and many others.
\par
Traditionally data is accumulated on the satellite and sent to an earth based antenna once it passes over ground stations \cite{ground_stations} which are located around the world or alternatively via a data relay satellite to the ground station. Then the data is transferred to where we need it on earth for processing and analysis. There has been ongoing research \cite{wertz2016next} into improving the downlink bandwidth such as switching from X-Band to Ka-Band \cite{shi2017development} but these improvements have been by a factor of 10 compared to the 200 times increase in data size. SAR image sensors are now capable of acquiring 2-5 TB of raw high resolution imagery per day which presents many opportunities but also huge challenges. According to Furano et al. \cite{furano2020ai} the capability that sensors have to produce data increases by a factor of 100 every generation, while our capabilities to download data are increasing, but only by a factor of three, four, five per generation. This fact combined with the predicted rise in the number of satellites is going to result in an exponential and possibly unsustainable rise in demand for the current communication infrastructure. 
\subsubsection{Satellite Image Processing} \label{Satellite_Image_processing}
Satellite image processing refers to the operation of extracting useful information from imagery acquired from a satellite. The processing steps are broken down as follows \cite{sowmya2017remote}

\begin{itemize}
\item Pre-processing - the operations required prior to the main data analysis to correct the image for any sensor irregularities and remove unwanted sensor distortion or noise. The categories of pre-processing include geometric correction, atmospheric correction and radiometric correction
\item  Image Enhancement - improving the image quality to a better and more understandable level for feature extraction or image interpretation. The categories of enhancement include radiometric, spatial, spectral and geometric
\item Image Transformation - generation of new images from current sources to highlight particular features or properties of interest. An example is to combine 2 images of the same area within different spectral bands or taken at different times. 
\item Image Classification and Analysis - the final step is using the outputs of the previous steps to isolate what we are looking for. 
A review of the various types of satellite image classification and analysis was done in 2015 \cite{abburu2015satellite} and also in 2019  \cite{babbar2019satellite} which included references to several data sets for evaluating algorithms and another more recently \cite{ouchra2021satellite}. They are subdivided between supervised (requiring a dataset containing labeled images of the object of interest used for training a model) and unsupervised (classifying an unlabeled dataset using techniques such as clustering to identify unlabeled classes/clusters)
The use of deep learning techniques \cite{krizhevsky2012imagenet} has revolutionised the area of image analysis, classification and segmentation \cite{hao2020brief}. Convolutional Neural Networks(CNNs) is now regarded as the most dominant approach when it comes to image classification. \cite{he2015deep}
\end{itemize}

\subsection{Artificial Intelligence}
Around the same time the 1st satellite was being launched, the domain of Artificial Intelligence (AI) was being created at a workshop at Dartmouth College \cite{mccarthy2006proposal}. Artificial intelligence techniques act as an enabler to extract value from data by identifying unknown patterns and objects. The concept of a machine replicating the human brain seemed like science fiction and despite a lot of investment in time and money it remained beyond the grasp of researchers for many years after. It wasn't until the late 20th century due to the improvement in computational hardware that AI started to show it's true potential with the use of machine learning (ML) models in headline applications such as autonomous vehicles, medical diagnosis and facial recognition. ML is a computationally demanding process and requires sufficient hardware to meet the demand. 

\subsection{Edge Computing}
Edge computing is the term given to the distributed computer paradigm where data processing occurs at or close to where the data is produced and/or consumed at the "edge" of the network.  While the term was originally coined back in the 1990's to describe the mechanism to deliver website content from a local server, it is more recently associated with the concept of Internet of Things (IoT), where devices embedded with sensors connect and exchange data with other devices or system connected to the internet. These IoT devices have grown in number at a enormous rate as they are now deployed in everyday consumer devices such as smart watches, home heating systems and fridges as well as vehicles and many healthcare devices to name but a few. The latest IoT Analytics report  \cite{StateofIoT} shows there are almost 15 billion connected IoT devices around the globe worth about \$800 billion and this is expected to grow to over 25 billion by 2030. These produce too much data to be handled by a cloud computing infrastructure which involves transporting the data over a network to a hosted processing service so the concept of having the service closer to the data was more applicable. The benefits include an improved response time for any applications that need to consume the output of the processing, while also reducing the demand on the network as the data doesn't have to travel so far. 
\newline There is a parallel between the problem of handling the data produced by the billions of IOT devices and the issue of handling the large volume of data being produced by EO satellites so there may be lessons to be learned from the edge computing domain which has been evolving over the last 20 years.

%% file: Sections/Constraints.tex
\section{Constraints to deployment of Machine Learning to Satellites} \label{ConstraintsondeployingMachineLearningtoSatellites}
\IEEEPARstart{T}{here} are multiple constraints that make it challenging to deploy machine learning (ML) models to EO satellites. We are focusing on deploying pretrained ML models so we are assuming the ML training is done on well resourced hardware pre-deployment.

\subsection{Power}  \label{Constraints_Power}
The only power source available to a satellite during it's lifetime is the power it gets from it's solar panels and stored in batteries on board. The design of the satellite power system has to balance power consumption with power generation as described by Lee et al.\cite{lee2013design}. The amount of power generated by a Low Earth Orbit (LEO) satellite is influenced by various factors (eg Solar panel size, efficiency \& satellites orbit) all of which are taken into account during the design phase. Some examples of satellites currently in orbit listed in the UCS Satellite Database \cite{ucs_satellite_database} are Landsat 8 launched in 2013 which generates 4.3KW or EMISat launched in 2022 which generates just 0.8W. Running a ML model on board will add to the power consumption demand in an environment where power is already very constrained and tightly managed so this will have to be fully understood to ensure the power generation can meet the increased demand.

\subsection{Processing Capability}  \label{Constraints_ProcessingCapability}
Processing capability refers to the ability of a computing system or device to execute computational tasks efficiently and effectively. It is a measure of the system's ability to perform calculations, process data, execute algorithms, and run software applications within a given timeframe.
Lentaris \cite{lentaris2018high} does a review of the typical standard space processing technologies and their applications. Routine housekeeping tasks such as attitude control,  command decoding, and systems management require very little processing power and can be supported by simple 8-bit microcontrollers(MCU), whereas more intensive tasks such as data compression, formatting or encryption require more advanced microprocessors. These tasks usually require a dedicated controller to interface with the platform. This will typically be either a microprocessor or a digital signal processor (DSP) device. 
\newline ML models typically involve a huge number (millions) of calculations, carried out by a processor which has to have sufficient computing capability to support these calculations. This is a large step up from basic tasks such as altitude control or data encryption and represents a large hurdle to be cleared to support deployment to a satellite. The processors that can meet this compute demand are typically a large GPU, multi core CPU or a combination of the 2 which take up significant space.  

\subsection{Memory} \label{Constraints_Memory}
The types of memory used on EO satellites are Volatile, Non-Volatile and Mass Memory. Each serve different purposes.
\smallskip
\newline The volatile memory serves the following purposes:
\begin{itemize}
    \item Program Execution: store the satellite's operating system, onboard software, and running processes. This includes the instructions and data needed to execute commands, perform computations, and control satellite subsystems.
    \item Data Buffering: serves as a temporary buffer for storing incoming data from onboard sensors or communication systems before it is processed, analyzed, or transmitted.
    \item Temporary Storage: store intermediate results and variables generated during onboard data processing tasks, such as image correction, compression, or feature extraction.
    \item System State: holds the current state of the satellite's subsystems, including sensor configurations, telemetry data, and system diagnostics
\end{itemize}
Common types of volatile memory used in EO satellites include Dynamic Random Access Memory (DRAM) which provides fast read and write access but requires continuous power to retain data and Static Random Access Memory (SRAM) which is faster and more power-efficient than DRAM and often used for critical subsystems where fast access times are essential. The size of volatile memory depends on factors such as the complexity of the satellite's systems and the processing requirements of onboard algorithms but typically ranges from several megabytes to gigabytes. Adding ML data processing will demand significant amount of volatile memory during inference and this is where the biggest constraint will be.
\smallskip
\newline Non-volatile memory serves the following purposes:
\begin{itemize}
    \item Permanent Storage: store essential software, firmware, operating system files, and configuration data required for the satellite's operation. 
    \item Data Logging: log telemetry data, sensor measurements, satellite health diagnostics, and other mission-critical information over extended periods. This data can be later retrieved and analyzed for mission planning, troubleshooting, and performance evaluation.
\end{itemize}
Common types of non-volatile memory include Flash memory (which offers fast read and write access, low power consumption, and resistance to shock and vibration) and Electrically Erasable Programmable Read-Only Memory(EEPROM) (used for storing small amounts of configuration data, calibration parameters, and persistent settings)
\smallskip
\newline Memory mass units (MMU) serves as a buffer for storing large volumes of data captured by the satellite's sensors. It allows the satellite to continue collecting data even when the ground station is not in range or when downlink bandwidth is limited. It stores various types of observational data, including images, spectral measurements, telemetry data, and sensor readings and usually organized into files or data packets for efficient storage and retrieval.
The amount of MMU memory required depends on factors such as the type and number of imaging sensors, the resolution (spatial - number of pixels in the image \& spectral - number of spectral bands), swath (area imaged on the surface) \& revisit time (the time between subsequent observations of the same area of interest). Typical mass memory sizes range from several gigabytes to terabytes.
It is provided in the form of Hard Disk Drives (HDD) and Solid State Drives (SSD) and is sometimes included as a module within the On Board Computer (OBC).  Any machine learning model deployed will be stored within the MMU so will have to be a consideration when sizing the MMU.

\subsection{Radiation} \label{Constraints_Radiation}
Space is one of the most extreme environments imaginable. During launch, the contents of a spacecraft will be subject to intense and violent shaking along with extremely loud sound waves. Once it gets to space it will them have to survive very high and low temperatures but probably the biggest risk to the functioning of any electronic devices is the radiation that exists in space. Maurer et al. \cite{maurer2008harsh} outlines the damaging impact that space radiation has on the electronics within space systems. 
Lange et al. \cite{lange2024machine} surveys the obustness of on-board ML models to Radiation

%% file: Sections/Solutions.tex
\section{Mitigating the Constraints to deployment of Machine Learning to Satellites} \label{Methodstomitigate}
\IEEEPARstart{T}{here} are several approaches to mitigating the constraints outlined in section \ref{ConstraintsondeployingMachineLearningtoSatellites} and these are outlined in this section.

 \subsection{Power}\label{Solutions_Power}
\begin{figure}[h!]
\includegraphics[scale=0.45]{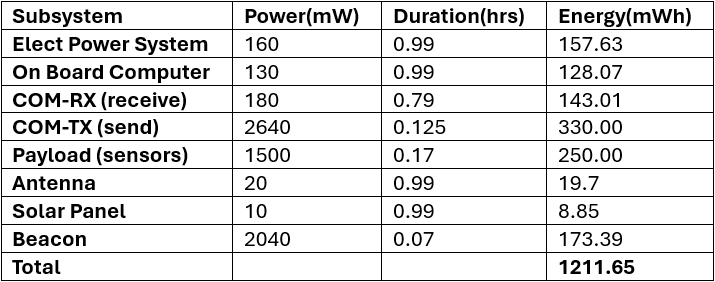}
\centering
\caption{Power consumption during day with transmission \& payload ( \cite{dahbi2017power}}
\label{tab:PowerConsumptionSatelliteSubsystems}
\end{figure}

During the design of a satellite, a power budget is produced to balance all the power required for each subsystem with the power generated by the solar panels. A good example of this is shown in Figure ~\ref{tab:PowerConsumptionSatelliteSubsystems} prodcued by Dahbi et al. \cite{dahbi2017power} who did a power budget for a  1U CubeSat orbiting the earth 15 times per day which equates to a 94.6 minute orbit (59mins / 0.99hr of which are in daylight) and has a power system that delivers 1568mWh per orbit. 

\begin{figure}[ht] 
\includegraphics[scale=0.85]{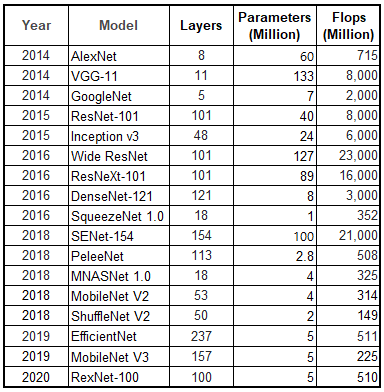}
\centering
\caption{Comparison of models FLOPS}
\label{fig:model_flops}
\end{figure}

\noindent Figure ~\ref{tab:PowerConsumptionSatelliteSubsystems} shows the energy consumption during daylight during which the satellite sensors (payload) are measuring and data is transmitted (COM-TX) data. This is the max consumption scenario but there's 15 other less consumption scenarios when the satellite may not be in daylight, measuring or transmitting data. This is all taken into account when balancing the power budget so that the total consumption does not exceed the power generated. If this is not the case then either the power system has to be modified to increase the power delivered or the power consumption reduced. This is generally done by either reducing how often the satellite payload operates or the number or times data is transmitted. In the example case the author decided the satellite would use the payload one orbit per day and communicate to the ground 4 orbits per day with the other 10 orbits used for generating power. 
\begin{figure}[ht]
\includegraphics[scale=0.75]{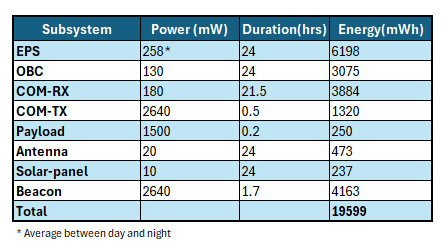}
\centering
\caption{Power consumption over 24 hours}
\label{tab:PowerConsumptionSatelliteSubsystemsOver24Hours}
\end{figure}
\noindent The cumulative consumption of each subsystem over the full day (24 hours) is shown in Table ~\ref{tab:PowerConsumptionSatelliteSubsystemsOver24Hours}. The biggest consumers over the day are the Electrical Power System (EPS), Beacon, the On board computer (OBC) and data transmission to (COM-TX) and from (COM-RX) the ground. \smallskip
\newline Using ML to process the data on board will consume a large amount of power. However a likely benefit to doing on board processing will be to reduce the volume of data that needs to be transmitted which directly correlates to a reduction in the transmission power consumption. Minimising the power consumed during data processing and maximising the reduction in data transmission consumption is the key to justifying the deployment of ML on board. 
There are various strategies to reducing the power consumption both at the hardware and the software level.  
\smallskip
\newline  Computer hardware systems technology has seen huge changes in recent years, well documented by Gill et al. \cite{gill2024modern}, driven by rapid technical and user-driven evolution, producing new paradigms such as cloud, fog \& edge computing. The proliferation of billions of embedded computing devices to every corner of the world has led to many challenges to enable computing in low power environments. 
Sze et al. \cite{sze2017hardware} explored how some of the challenges can be addressed at various levels of hardware design ranging from architecture, hardware-friendly algorithms, mixed-signal circuits, and advanced technologies (including memories and sensors). 
Power consumption is divided into two modes: active consumption takes place during data processing while static consumption takes place when data processing is not happening. Techniques such as  Dynamic voltage and frequency scaling (DVFS), Adaptive voltage scaling (AVS) \& Dynamic powerswitching (DPS) can save significant power within active mode while a technique known as static leakage management can  produce several very low power scenarios during the static mode. The author points out the efficiency benefits of performing MAC operations in parallel through the use of temporal architectures within both CPU and GPU. He also discusses the broader area of optimisation and energy efficiency of using accelerators. An AI accelerator is a high-performance parallel computation device that is specifically designed for the efficient processing of AI workloads. They draw much (100-1,000x) less power and are faster \& more scalable than general purpose devices. Reuther et al. \cite{reuther2022ai} did a survey of AI accelerators and processors plotting their performance and power values on a scatter graph. They span from extremely low power through embedded and autonomous applications to data center class accelerators for inference and training. Ortiz et al. \cite{ortiz2023onboard} investigated the use of accelerators to do onboard processing in satellite communications identifying AI-capable chipsets that include an On-Chip accelerators as well as a CPU that could be deployed on board a satellite. Low ($<$20W) powered options include the Intel/Movidius Myriad Family and several from the Nvidia Jetson family such as the Nano and TX2. The NVIDIA Jetson Nano is leading the way in enabling consumers to equip billions of low-power AI/ML systems across every domain and compares very favourably in terms of power consumption \cite{suzen2020benchmark}.
\smallskip
\newline On the software side there are a few approaches to reduce the power consumption. 
\newline Model Architecture: The design of the ML model used is made up of a combination of multiple layers. The unique combination of layers and layer parameters is called the model architecture and this has a direct impact into how much hardware resources are required to run the model. Many different model architectures have been developed since LeNet-5 which is regarded as a pioneering 7-level CNN was created in 1998. The aim of the model design was traditionally around maximising accuracy but more recently has included optimising the model, so it can be deployed to the available resources on the destination environment.
AlexNet \cite{krizhevsky2012imagenet} which won the ImageNet competition in 2012, was considered a revolutionary advancement outperforming other models by over 10\%. It consisted of five convolutional layers and three fully-connected layers with several innovations including: the ReLU activation function instead of the tanh function, use of Dropout which randomly drops neurons to avoid overfitting, a design which could use multiple GPUs in parallel \& the use of augmentation to create new data based on old data to also mitigate overfitting. 
Two years later the next big step improvement was by a group called the Visual Geometry Group (VGG) from Oxford University who created VGGNets \cite{simonyan2014very} which were a number of very deep CNN's, using smaller 3 × 3 convolution kernels (filters) rather than 5 × 5 and supporting up to 19 layers all of which use ReLU activation, and achieving even better results than AlexNet.
In the same year GoogleNet (or Inception-v1) introduced the idea of an Inception layer, running multiple convolutions in parallel. The number of layers was increased up to 22, but produced a large improvement in accuracy, pushing VGG into 2nd place for the 2014 ImageNet Classification competition. There were 4 different versions of Inception created with various architecture improvements before it was merged in 2016 with ResNet by Szegedy et al.\cite{szegedy2016inception}. 
ResNet had been introduced in 2015 by He et al. \cite{he2015deep} in a proposal to solve the Vanishing Gradient Problem associated with deep layer network training as outlined by Hochreiter \cite{hochreiter1998vanishing}. 
ResNet introduced the idea of \say{identity shortcut connection} that effectively skips layers. The authors demonstrated the approach by training a 1001 layer network to outperform less deep networks and it went on to win the ImageNet competition that year. It spawned several significant variants such as pre-activation \cite{he2015delving} described as \say{a robust initialization method that particularly considers the rectifier nonlinearities} which proved to be the first model to surpass the reported human-level performance on ImageNet classification (5.1\%) \cite{russakovsky2015imagenet}. 
In 2018 SENet \cite{hu2018squeeze} introduced another irregular module called Squeeze-and-excitation which improved the accuracy but at a huge computational cost of 100 million paramaters and over 10 billion operations. This was the last of the really deep and computationally expensive models with the focus since then being more on on designing lightweight and efficient networks for resource limited platforms such as IoT deployments, mobile devices or embedded devices. Squeezenet \cite{iandola2016squeezenet} was the one of the earliest models aimed at resource constrained devices employing design strategies to reduce the number of parameters while maintaining high accuracy.  In 2018, Tan et al. proposed Mnasnet \cite{tan2019mnasnet} generated from an automated neural architecture search (NAS) where the model latency is set as the main objective along with accuracy with the focus on deployment to mobile devices. MobileNets \cite{howard2017mobilenets} is a series of 3 lightweight models introduced by Google and was TensorFlow’s first mobile computer vision models. It uses depthwise separable convolutions and less layers and parameters compared to previous models.
The two ShuffleNet models \cite{zhang2018shufflenet} continued the trend focusing on mobile platforms with very limited computing power combining pointwise group convolution (ie 1x1 kernel that iterates through every single point) and channel shuffle within just 11 layers with accuracy comparable to MobileNet with 13× speedup compared to AlexNet. 
PeleeNet \cite{wang2018pelee} is a variation of DenseNet using regular convolutions instead of the less eficient depthwise convolutions. The table in Fig ~\ref{fig:model_flops} shows some of the well known models including several lightweight models along with their number of layers, parameters and flops on the PyTorch framework and trained on the ImageNet dataset.

\smallskip
\label{Optimisation}
\noindent Optimisation: Techniques that have been developed to adapt machine learning models to enable them to be deployed on hardware subject to resource constraints. Collectively they are called acceleration or optimisation techniques. The goal of the optimisation is to reduce the computational complexity during training/inference and hence allow the model to function on less resourced platforms.
Véstias \cite{vestias2019survey} investigated the high redundancy within CNN models suggesting scope for a significant simplification of the model. There have been many efforts to try and take advantage of this redundancy which broadly classify optimisation approaches under the following:
\newline
\begin{itemize}
    \item Layer decomposition - an approach to reduce the network computational
complexity by reducing the size of individual CNN layers.
    \item Knowledge Distillation - a training strategy of distilling and transferring knowledge from a large (teacher) model (or an ensemble/group of models) to a smaller (student) one to improve accuracy and speed
    \item Pruning - removing parameters deemed unnecessary
    \item Quantization - reducing the bitwidth of values by encoding full-precision parameters (ie weights and activations) with lower-precision
ones
\end{itemize}

Duggan et al. \cite{duggan2023profiling} investigated the power consumption behaviour of various ML algorithms deployed to a resource limited ‘edge’ devices such as a satellite and also the effectiveness of quantisation in reducing the power consumption during inference of up to 87\% without significantly affecting accuracy.

\subsection{Processing Capability} \label{Solutions_ProcessingCapability}
Advances in processor chips have improved the computing capability without impacting the size or power requirements and opening up satellite on board processing opportunities.
\par
\begin{figure*}[ht]
\includegraphics[width=\textwidth]{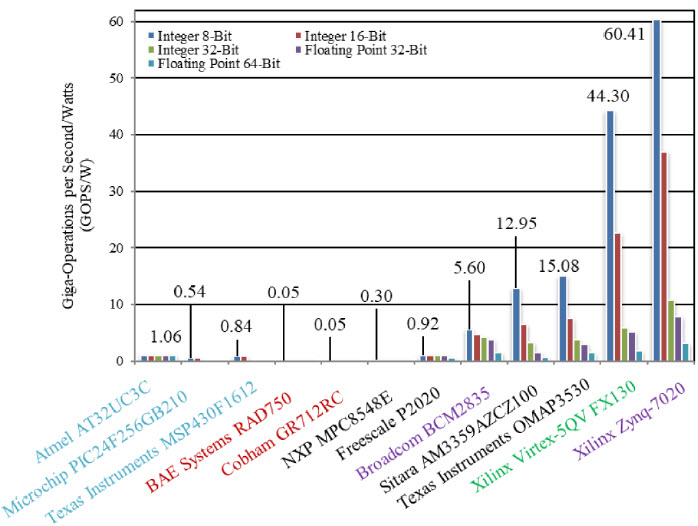}
\centering
\caption{Performance scaled by power comparison of onboard processors ( \cite{george2018onboard}}
\label{fig:Processor power comparison}
\end{figure*}

 In 2018 George et al. \cite{george2018onboard} investigated the challenges and opportunities for onboard computers within small satellites. Two possible approaches are proposed to meet the challenge:
\begin{itemize}
    \item Reconfigurable Computing - this is an alternative architecture to the expensive custom built hardware such as application-specific integrated circuits (ASICs). He proposed the use of field-programmable gate arrays (FPGAs) and lists some of the advantages of this approach including energy efficiency,the ability to load new software while the satellite is in orbit and parallelisation support.
    \item Hybrid Computing - a combination of several computing technologies to gain the advantages of each of them (eg combinning FPGA with ASIC), such as combining a radiation hardened device with a higher grade commercial device to achieve both high reliability and performance. An example is a System-on-chip (SoC) device used commonly on mobile systems.
\end{itemize} A good overview is given of the SmallSat Computing domain and the current range of processor options available from small microcontrollers to powerful microprocessors and the pros and cons of each. He provides a useful plot Figure \ref{fig:Processor power comparison} showing the performance normalized by power consumption of some of the common devices color coded by type: microcontrollers (light blue), radiation-hardened (rad), microprocessors (black), Field Programmable Gate Arrays(FPGA) (green), and System-on-Chip (SoC) (Purple). SoCs perform best with FPGA not far behind.

\par In 2017 Schmidt et al. introduced SpaceCubeX \cite{schmidt2017spacecubex}, a platform for the exploration of candidate hybrid
CPU/FPGA/DSP processing architectures. This led to the development of Space-
Cube v3.0 processor card \cite{geist2019spacecube} in 2019 by NASA, combining a a Kintex Ultra-Scale Field Programmable Gate Array (FPGA) and Xilinx Zynq Multiprocesser System on Chip (SOC) and considered an evolutionary advancement of spaceflight computing capability.
\par Nvidia launched their Jetson range of System-on-Chip (SoC) embedded computing boards back in 2014 as a low-power system designed for accelerating ML applications. They are increasingly becoming the market leader in this space and their small size \& high power to weight ratio makes them very suitable for deployment to a small satellite. 
\par One of the earliest efforts to host a computer vision ML algorithm on a simulated smallsat environment was performed in 2017 \cite{buonaiuto2017satellite}. It was an early proof of concept, using a Nvidia Tegra X1 (TX1). In 2018 Arechiga et al. \cite{arechiga2018onboard} demonstrated the use of of a Jetson TX2 using a CNN trained on satellite imagery from Planet’s Open California dataset. 
The same year \cite{manning2018machine} Manning explored the same domain to try and support intelligent computing on a spacecraft. Using a SoC Device (Xilinx Zynq-7020) he trained various CNN architectures on a dataset of 8000 terrestrial images and compared the accuracy (all over 90\%), execution time (89-1383ms) and runtime memory usage (8 - 41MB) with various parameters. The conclusion he drew was that it is possible to achieve acceptable performance with modern CNN models on a low-memory, low-power, space-grade, embedded platform.
In 2019 Yao et al.\cite{yao2019board} discussed and demonstrated (on ground) the use of the Jetson TX2, as a suitable embedded component for data processing (on board ship detection) on small satellites because of it's low cost, weight, size \& power consumption.
Again in 2019 Mittal \cite{mittal2019survey} carried out a survey of works that evaluate and optimize neural network applications on the Jetson platform, validating it as a viable hardware accelerator in a wide-range of application areas and comparing it against alternative platforms. The same year Slater et al. \cite{slater2020total} tested the Jetson Nano for radiation tolerance and concluded it could function successfully in a space environment for short term missions. 
Later the same year Hernandez et al. \cite{hernandez2019conceptual} carried out a feasibility study on the integration of a Jetson TX1 as the on-board processor for a CubeSat to deliver 1 TFLOP/s ( to calculate one trillion floating-point operations per second) of processing power. A suite of different algorithms were tested on the platform including objects recognition and classification using a neural network. The performance was evaluated in terms of how much quicker the TX1 performed the algorithms compared with a typical CubeSat microprocessor (ARM Cortex-A57 ) and quotes it as an acceleration factor. The results showed the TX1 performing over 14 times faster. He does however say that the peak power usage was 8.91W which may be too high depending on how much power is being generated by the CubeSat. 
\par In mid 2020 a very significant paper was published  \cite{giuffrida2020cloudscout} that led to the 1st real world deployment of Deep Learning on-board a SmallSat. CloudScout was introduced as a CNN for filtering images to be transmitted to ground operating directly on board a satellite. Since the paper was released the CNN has been installed on a satellite called HyperScout-2 and launched as part of the $\phi$-sat-1 mission, supported by the ESA through a program called PhiSat \cite{esa_phisat}. The key component is the Eyes of Things (EoT) board \cite{deniz2017eyes} powered by the Intel Movidius Myriad 2 Vision Processing Unit (VPU). An extracted dataset from the Sentinel-2 mission containing 21,546 hyperspectral cubes was used for training and testing the model. CloudScout running on Myriad 2 exceeded the requirements, performing an inference in only 325 ms with a model footprint of 2.1MB, accuracy of 92\% and a false positive of 1.03\% and an average power consumption of just 1.8 W. Crucially Myriad 2 also supports reconfigurability so it can be reprogrammed while in orbit. As a comparison a GPU CNN implementation consumed 100-200W and a CPU implementation consumes about 50W. \cite{li2016evaluating}.  The same author published a related paper \cite{giuffrida2021varphi} the following year to summarise the results of the deployment.

\par A benchmark comparison was done in April 2021, between a new proposed custom hardware accelerator designed for Field Programmable Gate Arrays (FPGAs) \cite{rapuano2021fpga} and the approach taken by CloudScout \cite{giuffrida2020cloudscout} using a Myriad 2 VPU. The motivation was to mitigate the fact that the Myriad is not yet considered suitable for the space environment. The results were that the FPGA-based solution had a reduced inference time of 142ms compared to 346ms for CloudScout, but at the cost of increase in power consumption of 3.4W compared to 1.8W for CloudScout, and a longer time to market as it is not off-the-shelf. As a conclusion, the proposed approach was considered more suitable to long-term Low Earth Orbit (LEO) satellites or interplanetary exploration missions who could benefit from the custom approach and would be more exposed to radiation and could utilise the radiation protection offered by FPGA's.

\par In mid 2022 NVIDIA launched the most recent of their ranges called the Orin marking a big leap forward in it's AI capabilities. The Jetson AGX Orin 32GB was released first, followed in early 2023 by the Orin NX and Orin Nano which has an 8GB variant containing a 1024-core GPU and 16 Tensor cores providing 40 Sparse or 20 Dense Tera-Operations per second (TOPs).

In late 2022 \cite{barnell2022ultra} Barnell used the Jetson Orin AGX hardware to run a deep learning object detector on a video feed from a drone detecting vehicles in real-time from various flight altitudes. In late 2023 Rad et al. \cite{rad2023preliminary} assessed the performance and power needs of both the Jetson Nano and the Jetson AGX Orin confirming the viability of using the NVIDIA Jetson devices for space applications involving demanding data processing or ML models. Around the same time Duggan et al. \cite{duggan2023profiling} profiled the power consumption of the Jetson Orin Nane 8GB platform classifying imagery using various convolutional neural networks (CNN) models.

\subsection{Memory} \label{Solutions_Memory}
As outlined in earlier section \ref{Constraints_Memory} there are various types of memory within the EO satellite serving various purposes. The addition of ML will increase the demand on both the memory mass unit (MMU) for storage of the ML algorithm and the Volatile memory (RAM) during execution of inference when the ML model is loaded into memory.
The approaches to mitigate memory as a constraint can be divided between increasing the storage capacity and reducing the demand for storage. 
\par Increasing the storage capacity is beyond the scope of this paper but it is worth noting that there is lots of new hardware offerings to the satellite development community to handle the increasing demand. Two of the more recent products which are built to support on board data processing are the Unibap SpaceCloud's iX10 \cite{unibap_spacecloud_ix10} which has 24GB RAM and 2 x 3.8TB SSD Storage and KP LAbs Antelope data processing unit (DPU) \cite{kplabs_antelope} which has 8GB RAM along with 4 GB SLC NAND Flash and an option of adding a further SATA SSD hard drive for storage. 

\par
\begin{figure}[ht]
\includegraphics[scale=0.85]{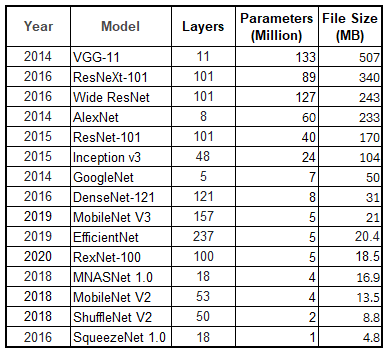}
\centering
\caption{Model Memory Requirements Comparison }
\label{table:model_file_size}
\end{figure}

 The amount of memory needed for ML model storage depends primarily on the complexity of the ML architecture, the number and precision of the model parameters along with the training dataset. The ML frameworks (e.g., TensorFlow, PyTorch) use different serialization formats with to save trained models and this can also affect the model file size. There are various different optimisation and compression techniques which can greatly reduce the file size. Model sizes can vary from a few kilobytes to hundreds of megabytes depending on these factors. Figure \ref{table:model_file_size} shows some of the more common models built on the PyTorch framework and trained on the ImageNet dataset. The largest is the VGG models at over 500MB and the smallest is SqueezeNet less than 1\% of that size at 4.8MB.

\smallskip
Some model optimisation techniques (Layer decomposition, Knowledge Distillation,Pruning \& Quantization ) were detailed in the previous power reduction section \ref{Optimisation}. Along with reducing the power required to run these models they can also reduces the memory footprint of a model as well as the memory required to run the model, making it more feasible to store and use on devices with limited memory capacity.  
\par Pre-processing describes the operations required prior to the data analysis to correct the image for any irregularities and generally done for improving the accuracy of the data analysis. Using ML models offers the opportunity to do some more advanced pre processing reducing the memory consumed. A good example of this is the Cloudscout segmentation neural network (NN), run on Myriad 2 as part of the $\phi$-sat-1 mission and described by Giufrfrida et al. \cite{giuffrida2020cloudscout} to identify, classify, and eventually discard on-board the cloudy images. Similarly Maskey \cite{maskey2020cubesatnet} as part of the BIRDS-3 project \cite{birds_3_project_2021} used a a lightweight CNN architecture called CubeSatNet to filter images before transmitting and also  Zhang et al. \cite{zhang2018cubesat} who introduced a new framework for detecting clouds from on-board a satellite saving 66\% storage content. According to the Irish Meteorological Service, the sky above Ireland is entirely cloudy over 50 percent of the time with more clouds during the day than at night, and fog also common. So filtering these images is a very significant memory saving. 
Another case is Arechiga et al. \cite{arechiga2018onboard} who introduced a 2-step approach for ship detection from on-board a SmallSat. Neither of these state explicitly how much memory is saved but we can reasonably deduce the savings is proportional to the amount of processing done. If all the data processing is done on board then only the output of the processing (eg location and details of an active forest fire or oil spill) needs to be transmitted to earth and none of the actual images may need to be sent. If partial processing is done (eg output of a specific neural network intermediate layer) then we can deduce this will be significantly smaller than the original image. This output could potentially be transmitted to Earth and the remaining processing done on Earth. Unfortunately the author could not find a reference paper with such work done so there is scope for further research in this area.

\subsection{Radiation} \label{Solutions_Radiation}
 Maurer et al. \cite{maurer2008harsh} describes the mitigation strategies to get hardware to be "radiation hardened", which is a requirement in many cases before electronic devices can be approved for deployment to space. The time delay caused by this process often results in approved hardware being at least one generation behind the latest hardware being used on earth.  

%% file: Sections/Conclusion.tex
\section{Conclusion and Future Research }
Earth observation satellites have seen a surge in numbers in the last decade due to a series of technological advancements and a significant reduction in costs. This has led to opportunities to exploit the data in all kinds of applications from disaster management to marine and land analysis. The increase in the volume of data produced has pushed the traditional satellite data workflow to its limits, and new approaches are being devised to handle the data, including the use of artificial intelligence models to process the data on board. This is a challenging task with several resource constraints to overcome. This paper reviews the most significant of these including the availability of sufficient power, processing capability and memory on board the satellite to facilitate the operation of these models. 
\newline A lot of research has been done into strategies to mitigate these constraints including selecting energy efficient components and machine learning models along with creating and balancing an accurate power budget for all the energy consumers within the satellite. Low power processor chips with excellent computing capabilities have been created and embedded onto systems small enough to fit on-board CubeSats. There have been several of these used on board small satellites to validate that they function well. The limited memory on board can be mitigated by choosing a model such as SqueezeNet which requires under 5MB. Also applying one of a number of well tested optimisation techniques can drastically reduce the momeory footprint without significantly impacting the performance. Pre-processing the imagery on-board can also reduce the amount of memory allocated for image storage.
\newline  The number of real-world examples of utilizing machine learning models for on-board satellite image processing remains relatively limited, highlighting the need for further research in this area. Minimizing the memory footprint of data transmitted from satellites to Earth offers substantial advantages, including reduced on-board memory requirements and lower transmission power consumption. The resulting power savings can compensate for, and potentially surpass, the energy needed to perform the processing. More research is required to quantify both the power and memory saved. Partial image processing is also an area that could be researched further to be able to quantify the resource savings possible. Addressing the resource limitations within a satellite to enable machine learning inference is a challenging task. However, it holds the potential to unlock numerous opportunities for leveraging the underutilized resource of satellite data, providing real-time, actionable insights.  